\documentclass[runningheads]{llncs}

\usepackage{graphicx}
\usepackage{hyperref}

\usepackage{placeins}

\let\Oldsection\section
\renewcommand{\section}{\FloatBarrier\Oldsection}

\let\Oldsubsection\subsection
\renewcommand{\subsection}{\FloatBarrier\Oldsubsection}

\let\Oldsubsubsection\subsubsection
\renewcommand{\subsubsection}{\FloatBarrier\Oldsubsubsection}

\begin{document}
\title{A Deep Dive into Perturbations\protect\\as Evaluation Technique for Time Series XAI}
\titlerunning{A Deep Dive into TS-XAI Perturbations}
%
\author{Udo Schlegel\inst{1}\orcidID{0000-0002-8266-0162} \and
Daniel A. Keim\inst{1}\orcidID{0000-0001-7966-9740}}
\authorrunning{Schlegel \& Keim}
%
\institute{University of Konstanz, Universitätsstraße 10, 78464 Konstanz, Germany
\email{\{u.schlegel,daniel.keim\}@uni-konstanz.de}}
\maketitle              
\begin{abstract}
Explainable Artificial Intelligence (XAI) has gained significant attention recently as the demand for transparency and interpretability of machine learning models has increased. 
In particular, XAI for time series data has become increasingly important in finance, healthcare, and climate science. 
However, evaluating the quality of explanations, such as attributions provided by XAI techniques, remains challenging. 
This paper provides an in-depth analysis of using perturbations to evaluate attributions extracted from time series models.
A perturbation analysis involves systematically modifying the input data and evaluating the impact on the attributions generated by the XAI method. 
We apply this approach to several state-of-the-art XAI techniques and evaluate their performance on three time series classification datasets. 
Our results demonstrate that the perturbation analysis approach can effectively evaluate the quality of attributions and provide insights into the strengths and limitations of XAI techniques.
Such an approach can guide the selection of XAI methods for time series data, e.g., focusing on return time rather than precision, and  facilitate the development of more reliable and interpretable machine learning models for time series analysis.

\keywords{Explainable AI  \and XAI Evaluation \and XAI for Time Series.}
\end{abstract}
\section{Introduction}
Artificial intelligence (AI) has become an integral part of our daily lives, from the personalized advertisement we receive on social media to conversational AI (chatbots) answering questions of users and customers using deep neural networks. 
However, as the complexity of deep neural network models increases, so does the difficulty in understanding how they get to their decisions~\cite{guidotti_survey_2018}. 
A lack of interpretability can lead to severe consequences in critical domains such as finance, healthcare, and transportation, including financial losses, medical errors, and even loss of life by providing wrong decisions if complex models are deployed~\cite{rudin_stop_2019}.
One promising approach to addressing such issues is through the usage of explainable artificial intelligence (XAI), which seeks to provide insights into the inner workings of complex models and the factors that drive their decision-making~\cite{guidotti_survey_2018}. 
One particular area of interest is time series data, which is characterized by the sequential nature of its observations and the interdependencies between them, as more sensors generate a massive amount of data and more tasks are tackled by complex models~\cite{theissler_explainable_2022}.

In recent years, a growing body of research has focused on developing XAI techniques tailored explicitly for time series data~\cite{theissler_explainable_2022}. 
These techniques often rely on the concept of attributions, which aim to identify the contributions of individual features and time points to the overall prediction made by a model~\cite{theissler_explainable_2022}. 
By providing insights into which parts of the input data are most relevant to the output, attributions can help users understand the reasoning behind the model's decision-making process~\cite{schlegel_time_2021}.
However, the evaluation of such attributions is not trivial~\cite{schlegel_empirical_2020}.
To address the challenge of evaluating the quality of explanations for time series data, perturbation analysis has emerged as a promising evaluation technique~\cite{schlegel_towards_2019,simic_perturbation_2022}. 
Perturbation analysis involves systematically modifying the input data and assessing the impact on the attributions generated by XAI methods~\cite{schlegel_empirical_2020}. 
By perturbing the input data, it is possible to evaluate the robustness of the explanations provided by XAI methods~\cite{theissler_explainable_2022}. 
However, the effectiveness of perturbation analysis for evaluating the quality of attributions for time series data has not been extensively studied~\cite{schlegel_empirical_2020}.

In this paper, we apply attribution techniques from various fields to a convolution neural network trained on time series classification data to evaluate and inspect the generated attributions in detail using perturbations, which involves systematically altering the input data and observing the effect on the model's output.
We investigate the performance of attribution techniques compared to each other based on the perturbation analysis result and explore the perturbation changes based on these attributions to gain insights into the model.
Through such an analysis, we can identify spurious correlations and shortcuts in the complex models and thus enable developers to potentially improve models by debugging datasets.
We show that our convolution neural network trained on time series classification learned certain shortcuts to achieve state-of-the-art performances.
Based on these experiments and results, we provide guidelines for the application of attribution techniques for time series classification and release our evaluation framework to investigate other attribution techniques. 

Thus, we contribute:
(1)~~an in-depth analysis of attribution techniques on time series classification for deep learning models using a perturbation analysis,
(2)~~insights into convolution neural networks trained on time series based on the generated attributions,
(3)~~guidelines and a framework for applying attribution techniques for time series models with a perturbation analysis card for reporting.
We first look into related work, and then we introduce the perturbation analysis methodology and the experiment setup we use for our deep dive.
Here we also propose perturbation analysis cards as a guideline to report the results of an evaluation.
Next, we present our results and discuss the impact of our conclusions for attribution techniques applied to time series.
Lastly, in future work, we motivate new measures for the evaluation of attributions on time series data.

Results and source code of the experiments is online available at:\\{\small\href{https://github.com/visual-xai-for-time-series/time-series-xai-perturbation-analysis}{https://github.com/visual-xai-for-time-series/time-series-xai-perturbation-analysis}}

\section{Related Work}

Explainable AI (XAI) accelerated through several surveys~\cite{guidotti_survey_2018,adadi_peeking_2018} and techniques, e.g., LIME~\cite{ribeiro_why_2016} and SHAP~\cite{lundberg_unified_2017} in the last few years.
Especially, attributions are prevalent in the image domain as heatmap explanations are easy to understand for users~\cite{jeyakumar_how_2020}.
Some theoretical works dig into the backgrounds of why models learn certain shortcuts to solve tasks~\cite{geirhos_shortcut_2020} and thus enable further explanations for decisions.
However, evaluating explanations is still a slowly growing area with limited work toward benchmarking different techniques against each other~\cite{hooker_benchmark_2019}.
Further, shortcuts or spurious correlations are not trivial to detect in explanations and need an in-depth analysis to be able to identify these~\cite{zhou_do_2022}.

Some works started to collect possible evaluation techniques~\cite{mohseni_multidisciplinary_2021} and categorized these into five measurements: mental model, explanation usefulness and satisfaction, user trust and reliance, human-AI task performance, and computational measures.
The first few measures focus on evaluating with or in cooperation with humans and are thus heavily influenced by human factors.
The computational measures exclude human factors and focus on purely automatic evaluation of explanations.
In this work, we inspect the computational measures and, more precisely, the explainer fidelity of the attribution technique on the model to show how the attributions fit the model.

XAI for time series classification (TSC), on the one hand, incorporates previously proposed explanation techniques from other fields and introduces the time dependence into some of the techniques~\cite{theissler_explainable_2022}.
Theissler et al.~\cite{theissler_explainable_2022} categorize possible explanations for TSC into time point, subsequence, and instance explanations.
All these operate on a different level of the time series and are thus unique in their explanation and evaluation.
In this work, we tackle time point explanations and, to be more precise, attributions to highlight and explore shortcuts and spurious correlations.
As Schlegel et al.~\cite{schlegel_towards_2019} and others~\cite{simic_perturbation_2022,mercier_time_2022,theissler_explainable_2022} demonstrated, attributions techniques such as LIME~\cite{ribeiro_why_2016}, SHAP~\cite{lundberg_unified_2017}, LRP~\cite{bach_pixel_2015}, GradCAM~\cite{selvaraju_grad_2017}, Integrated Gradients~\cite{sundararajan_axiomatic_2017}, and more~\cite{schlegel_ts_2021}, produce working attributions on time series to extract explanations from a model.
However, in most cases, only purely computational measures are applied to the attributions, which are not further inspected, e.g., Mercier et al.~\cite{mercier_time_2022} to gain deeper insights.

Schlegel et al.~\cite{schlegel_towards_2019} started by using a perturbation analysis on attribution techniques applied to TSC using various perturbation functions to highlight that techniques for images and text are also working on time series.
Based on such preliminary experiments, they enhanced their approach with additional perturbation functions to showcase deeper insights into the fidelity evaluation~\cite{schlegel_empirical_2020}.
Mercier et al.~\cite{mercier_time_2022} enhanced these perturbations with further measures from the image domain, such as (in)fidelity and sensitivity~\cite{yeh_in_2019}.
Simic et al.~\cite{simic_perturbation_2022} extended the proposed methods by Schlegel et al.~\cite{schlegel_empirical_2020} with out-of-distribution detecting functions and gave guidelines for the selection of attribution techniques and the size of the window for the perturbation.
Turbe et al.~\cite{turbe_interprettime_2022} enhance previous approaches with another metric to improve the comparison of the attribution techniques and the ability to demonstrate their fidelity towards the model.
However, all of these approaches do not look into the attributions and the produced values to investigate further into the techniques behind the attributions and the models.
Thus, an in-depth analysis is needed to investigate the attributions generated for time series classification models.

\section{Perturbation Analysis}

We use the perturbation analysis approach by Schlegel et al.~\cite{schlegel_empirical_2020} to generate attributions, verify, and compare them using the proposed perturbation function strategies~\cite{schlegel_towards_2019,simic_perturbation_2022}.
We extend the comparison by calculating the Euclidean and cosine distance between the original and the perturbed time series instance and the Euclidean and cosine distance between the original attributions of the dataset and the attributions of the perturbed instances of the dataset.
Collecting these results can help us have a more in-depth analysis of the attribution techniques and reveal relations between attributions and models.
However, we first need to establish the general perturbation analysis.

Let $D = (X, Y)$ be a time series classification dataset with $X$ as the time series samples and $Y$ as the time series labels.
$X = \{ts_1, ts_2, ..., ts_n\}$ contains $n$ time series samples with $m$ time points for each sample represented as $ts = \{tp_1, tp_2, ..., tp_m\}$, where $tp_1$ is the value for the $i$th time point of $ts$.
$Y = \{l_1, l_2, ..., l_n\}$ contains $n$ labels one label for each time series sample.
Let $M(ts, \theta) = y'$ be a time series classification model which predicts a label $y'$ based on a time series input $ts$ and has the parameters $\theta$.
Let $A(X, M, \theta)$ be an XAI technique for generating attributions for the time series data.
The original attributions for $X$ generated by $A$ can be represented as $A(X, M, \theta) = \{a_1, a_2, ..., a_m\}$, where $a_i$ is the attribution score for the $i$th time point of $X$, $M$ the time series classification model for which the attributions are calculated, and $\theta$ the parameters of the attribution technique.

\begin{figure}[h!tb]
    \centering
    \includegraphics[width=\textwidth,trim={0 177mm 225mm 0},clip]{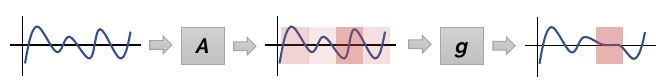}
    \caption{Starting from a time series $ts$, we use a selected attribution technique $A$ to get attributions. Based on the attributions, we use a selected perturbation function $g$ to set highly relevant time points, e.g., to zero. Further information in Schlegel et al.~\cite{schlegel_empirical_2020}.}
    \label{fig:pertrubation_analysis}
\end{figure}

To perform perturbation analysis, we introduce a perturbation function $g$ that modifies $X$ in a controlled manner. 
Specifically, we define a perturbed time series dataset $X'$ as:
\begin{equation}
    X' = g(X, A, \xi)
\end{equation}
Our perturbation function $g$ modifies the dataset $X$ based on the attributions $A$ and a threshold $\xi$.
The value for the modification can be changed and depends on the selected function $g$, e.g., exchange to zero.
The threshold $\xi$ can be set to a value by hand or some other function, e.g., using the 90-percentile of the attributions so that the attributions, e.g., $a_i$ the $i$th element, above the threshold, will be modified to the previously set value, e.g., zero.
\autoref{fig:pertrubation_analysis} demonstrates the approach with zero perturbations on attributions with high values.

The original $X$ and the perturbed dataset $X'$ get predicted with the model $M$ to get $M(X) = Y'$ and $M(X') = Y''$.
Based on Schlegel et al.~\cite{schlegel_empirical_2020}, we incorporate a quality metric $qm$, e.g., accuracy, to compare the performance of the model $M$ with the original $X$ and the perturbed dataset $X'$.
For the time series classification, we assume that the $qm$ decreases after the original data changes, and thus the labels are not fitting anymore~\cite{schlegel_towards_2019}.
We further assume a suiting attribution technique decreases the performance more heavily as the most relevant parts of the input data get perturbed~\cite{hooker_benchmark_2019}.
Thus, we assume:
\begin{equation}
    qm(Y', Y)\leq qm(Y'', Y)
\end{equation}

However, in some cases, the scores are very similar~\cite{schlegel_towards_2019,mercier_time_2022}, and a deeper investigation into the attributions is necessary to find similarities or dissimilarities in the relevances of the techniques.
Thus, we do not only compare the quality metrics but also focus on the distances between the original $X$ and the perturbed $X'$ datasets.
We apply the Euclidean and cosine distances to the datasets as these are common distance functions for time series~\cite{aghabozorgi_time_2015} to collect the changes of the perturbation function $g$.
We define the Euclidean distance as:
\begin{equation}
    Euc(X, X') = \sqrt{\sum_{i=1}^{n} (ts_i - ts_i')^2}
\end{equation}
where $X = {ts_1, ts_2, ..., ts_n}$ and $X' = {ts_1', ts_2', ..., ts_n'}$ are the two time series being compared.
And we define the cosine distance as:
\begin{equation}
    Cos(X, X') =  1 - \frac{\sum_{i=1}^{n} ts_i \times ts_i'}{\sqrt{\sum_{i=1}^{n} ts_i^2} \times \sqrt{\sum_{i=1}^{n} ts_i'^2}} 
\end{equation}
where $X = {ts_1, ts_2, ..., ts_n}$ and $X' = {ts_1', ts_2', ..., ts_n'}$ are the two time series being compared.
These changes enable us to compare the attributions not only on a performance level but on a raw level directly on the data.

\section{Experiments with Perturbation Analysis}

For our analysis, we look into the time series that changed and those that did not change during the perturbation analysis.
We especially want to understand the attribution distributions to investigate the attribution techniques responsible for fitting explanations, with high fidelity~\cite{mohseni_multidisciplinary_2021}, on the models.
Fitting explanations in our assumptions are techniques that change the prediction of more samples in a perturbation analysis~\cite{schlegel_towards_2019,schlegel_empirical_2020,mercier_time_2022}.
However, a general measure and metric for evaluating explanations are essential, but another factor is the attributions, as these can also hide information or present spurious correlations~\cite{zhou_do_2022}.
E.g., the question of how attributions are distributed over the techniques arises.

To answer such questions and others, we use the changes from $Y$ (old prediction) to $Y'$ (new prediction) to look into the samples that changed their prediction and those that do not change.
We especially want to know when a change in the prediction happened, e.g., after how many removals based on the attributions and the perturbation strategy.
Thus, we look at the prediction changes from one class to the other.
E.g., in a binary classification with the assumption from above, the predictions change from one to the other class to demonstrate that the attributions highlight relevant time points for the model.
Thus, we slowly perturb more and more values from the time series until there is a change in prediction.
We use the percentile values (99, 98, ..., 25) as a threshold for the perturbation and record when the change happens.
Further, we collect the skewness of the attributions of the changed and unchanged predictions.
With such an exploration of the distributions of the attributions, we enable to inspect patterns inside of the attributions generated by different techniques.
Also, the distributions of the skewness enable to have another factor for the comparison of the attribution techniques.
Lastly, we do not only collect the skewness but also the Euclidean and the cosine distances of the original sample to the perturbed instance with regard to the changed and unchanged predictions.
All these different collected statistics and properties can help us to identify various findings, insights, and correlations in the attribution techniques as we collect as much data from our perturbation analysis as possible.

\begin{figure}[h!tb]
    \centering
    \includegraphics[width=\textwidth,trim={0 21mm 0 1mm},clip]{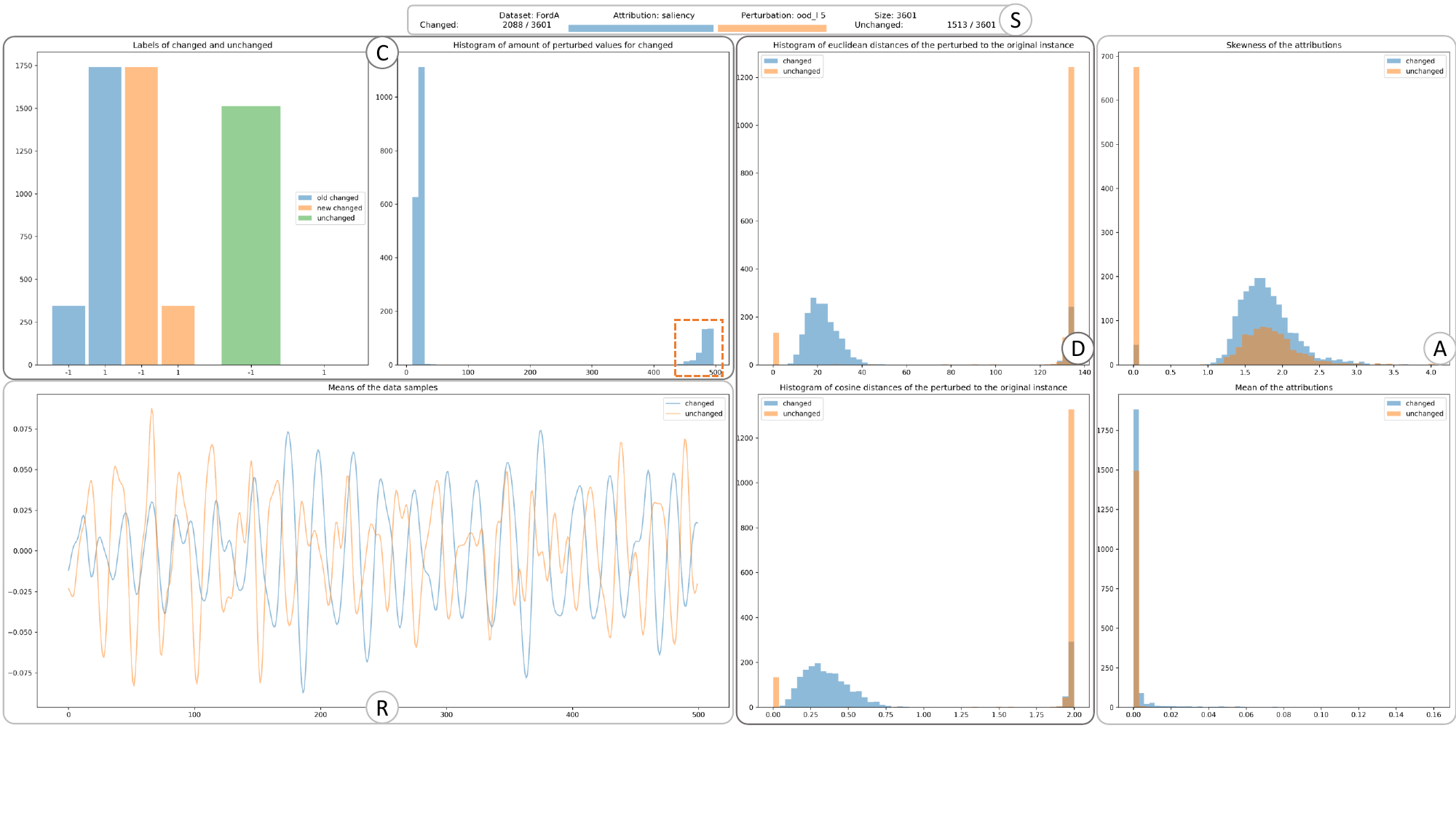}
    \caption{Perturbation analysis card for the FordA dataset: the top description (S) contains general statistics for the card, starting with the dataset, the attribution technique, and the perturbation strategy. Beneath are the statistics for the amount of changed and unchanged sample predictions encoded as a progress bar and with numbers. The plots in (C) give a more detailed insight into the class changes after the perturbation. The left plot presents the amount of changed and unchanged samples for each class and also visualizes the class change for changed predictions. The right plot shows the number of perturbed values when a change in prediction happens. In (D), the distances of the original to the perturbed instance are shown. The top presents the Euclidean distance between the pairs, and the bottom shows the cosine distance. (A) presents the skew (top) and mean (bottom) of the attributions for the changed and unchanged sample predictions. In (R), the mean of every value at a specific time point is visualized for the changed and unchanged samples. }
    \label{fig:perturbation_card}
    \vspace{-1em}
\end{figure}

\textbf{Summary -- }
Overall, we have the following dimensions we want to experiment on:
a) attribution techniques,  
b) perturbation strategy.
We collect and analyze the following properties:
a) mean of the raw data samples of the changed and unchanged predictions;
b) skewness of attributions based on the changed and unchanged predictions after the perturbation;
c) new class distributions of the changed and unchanged predictions after the perturbation;
d) amount of relevant attributions needed to perturb an instance to another class prediction.
\autoref{fig:perturbation_card} presents the collected properties using a perturbation analysis card with various statistics aggregated and visualized for easier analysis.
We created these perturbation cards for all the experiments.

\textbf{Hypotheses -- }
After we established our experiment setup, we generated hypotheses around the results of the experiment on the basis of other work.
Based on the preliminary experiments by Schlegel et al.~\cite{schlegel_towards_2019}, we generated the hypothesis that SHAP or SHAP derivatives will lead to the best results for the TSC task.
Based on the results of Simic et al.~\cite{simic_perturbation_2022}, we will further look into the other attributions and double-check the results of Schlegel et al.~\cite{schlegel_towards_2019} and the SHAP results even if SHAP results are less consistent~\cite{simic_perturbation_2022}.
Based on Simic et al.~\cite{simic_perturbation_2022}, we further look into the different perturbation strategies as we hypothesize that using one strategy is not enough to find a suitable attribution technique.
Based on Geirhos et al.~\cite{geirhos_shortcut_2020}, we want also to check if there are patterns in the data the attributions show as relevant to find shortcuts the model learned to classify the data.
E.g., using certain maximum or minimum values to separate one class from the other in a binary classification problem.

\textbf{Perturbation Analysis Card -- }
The perturbation analysis card is our proposed approach to reporting the results of our perturbation analysis strategies and techniques.
\autoref{fig:perturbation_card} shows such a perturbation analysis card with meta information (S), bar charts about the switch from one class to another (C), bar charts for the distribution of distances (D), statistics about the attributions (A), and a line plot visualization about the raw time series data (R).

Starting on top,~\autoref{fig:perturbation_card} (S), a short introduction presents a description of the dataset, the attribution technique, and the perturbation strategy.
Right under the description, a stacked vertical bar chart shows a short glimpse of how good or bad the overall perturbation was.
A good perturbation with an attribution technique presents just a lot of blue in this bar chart, while a bad perturbation shows a lot of orange in the visualization.
Next to it, the exact numbers of the changed and unchanged samples are shown so that comparable numbers enhance the fast glance with other cards.

\autoref{fig:perturbation_card} (C) gives a detailed view of the perturbation and the changes there.
The bar chart on the left visualizes the classes of the changed and unchanged predictions.
For the changed prediction, the visualization also further presents the classes before and after the perturbation.
Such visualization can help to identify spurious correlations as a model could, for instance, learn one feature of one class for the prediction.
The bar chart on the right at (C) shows the number of perturbed values needed to change the prediction.
The fewer changes needed, the better the attribution can identify relevant values.

In~\autoref{fig:perturbation_card} (D), the histogram of the distances between the perturbed and the original instances are shown.
On top of (D), the Euclidean distances, and on the bottom of (D), the cosine distance can help to find clusters of needed changes for the perturbation of the samples by revealing a trend towards a certain distance range.
Also, the distances can be used to compare the individual attribution techniques against each other.
A smaller distance range, together with a lower number of perturbed values, presents a more focused technique. 

\autoref{fig:perturbation_card} (A) visualizes more statistical information about the attributions.
The plot on top of (A) shows the skewness of the attributions of the samples of the dataset.
On the bottom, the means of the attributions are visualized.
Through these, a general trend of the changed and unchanged samples and their attributions can be seen.
Especially, outliers are interesting as a starting point for deeper analysis with other methods and visualizations.

Lastly, in~\autoref{fig:perturbation_card} (R), the time series time point means of the changed and unchanged samples can be inspected.
So, for every time point in the time series, the mean of it over the subset (changed or unchanged) of the whole dataset is calculated and visualized.
Thus, in the case of, e.g., FordA, with a standardization of the dataset, the samples slowly converge to zero.
The visualization enables to spot large differences between the changed and unchanged samples.

\section{Results and Discussion}

Our current experiment setup evolves around an in-depth analysis of the attributions of seven attribution techniques (Saliency, IntegratedGradients, DeepLift, Occlusion, GradientShap, DeepLiftShap, KernelShap) based on the implementations in Captum~\footnote{Captum is a Pytorch-based XAI module for Python: \url{https://captum.ai/}}.
We incorporate 16 perturbation strategies, two based on Simic et al.~\cite{simic_perturbation_2022}, six based on Schlegel et al.~\cite{schlegel_empirical_2020}, and eight extensions we describe later.
We implemented nine single time point perturbations (zero, mean, inverse, dataset mean, dataset max, dataset min, OOD high, OOD low, random between min and max) and seven subsequence perturbations (zero, subsequence mean, dataset mean, inverse, OOD high, OOD low, random between min and max).
The subsequence length is fixed to ten percent of the length of the data.

We focus on the UCR benchmark datasets~\cite{dau_ucr_2019} and take three of the most extensive datasets (FordA, FordB, ElectricDevices) to investigate data characteristics.
However, our approach can be applied to any time series classification dataset.
The FordA and FordB are sensor data with a length of 500 and provide an anomaly detection binary classification task.
FordA has 3601 samples in the training set and 1320 in the test set.
FordB has 3636 samples in the training set and 810 in the test set.
The ElectricDevices dataset is shorter, with only 96 time points.
However, the dataset has 8926 training samples and 7711 test samples.

We investigate two architectures of convolutional neural networks.
The first architecture tackles the FordA and FordB datasets.
The model consists of three 1D convolutional layers with a kernel size of three and increases the channels from one to 10 to 50 to 100.
A max-pooling of three after the convolutional layer decreases the size again.
A ReLu activation function activates the neuron.
Afterward, a fully connected layer with 100 neurons and a ReLu function uses the feature maps from the convolutional layers to process the data further.
And lastly, another fully connected layer with two neurons classifies the data with a softmax activation on top.
We train the model with a batch size of 120 and the Adam optimizer~\cite{kingma_adam_2014}.
The second architecture is trained on the ElectricDevices data and introduces a residual from the input to the fully connected layers.
The original input gets downsampled using a 1D convolution with kernel size seven for the residual addition right before the fully connected layers.

We train our models using the cross-entropy loss for multi-label classification on the datasets for 500 epochs.
Our models achieve for FordA an accuracy of 0.99 for the training set and 0.89 for the test set, demonstrating overfitting to the training data.
Our models achieve for FordB an accuracy of 0.99 for the training set and 0.70 for the test set, demonstrating overfitting to the training data.
Our models achieve for ElectricDevices an accuracy of 0.94 for the training set and 0.64 for the test set, demonstrating overfitting to the training data.
As Ismail Fawaz et al.~\cite{ismailfawaz_deep_2019} showed, even our simple models are not too far from the state-of-the-art with other more sophisticated models.
However, as we want to analyze our model, we look into the attributions of the training data, and thus our overfitting is a nice bonus to investigate spurious correlations and shortcuts~\cite{geirhos_shortcut_2020}.

\begin{figure}[!htb]
    \centering
    \includegraphics[width=0.99\textwidth,trim={0 43mm 77mm 0},clip]{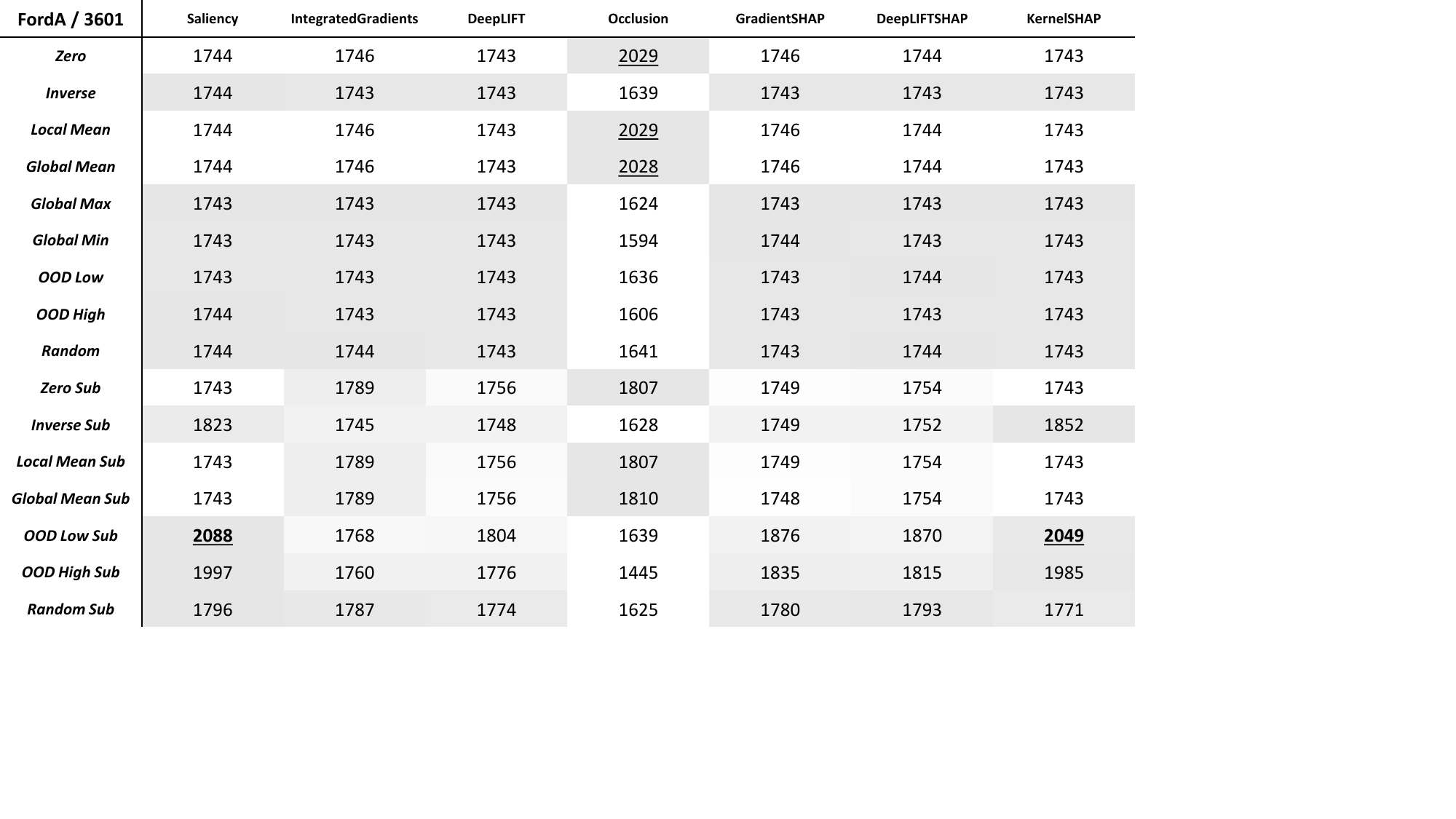}
    \caption{Changed samples from the perturbation analysis for the FordA dataset. The higher, the better. Saliency and KernelSHAP as winners, with Occlusion behind.}
    \label{fig:forda_results}
\end{figure}

\textbf{Results --}
First, we start with the FordA dataset; next, we will present the FordB results, and lastly, the ElectricDevices dataset.
FordA demonstrates interesting results regarding the attribution techniques and the perturbation strategies.
The best working strategies are setting the perturbed value to an out-of-distribution low~\cite{turbe_interprettime_2022} on a subsequence~\cite{schlegel_empirical_2020} as you can see in~\autoref{fig:forda_results}.
Especially, the saliency method~\cite{simonyan_deep_2014} achieves the best result regarding the flip of predictions by flipping 2088 of 3601 samples, as also seen in~\autoref{fig:perturbation_card}.
However, the KernelSHAP method~\cite{lundberg_unified_2017} comes close to the flip with just 39 less with 2049 flips.
Also, as seen in~\autoref{fig:perturbation_card} on the plot on the right, the perturbation strategy out-of-distribution low changes the class quite late with a lot of perturbed values.
Such an effect is unwanted in many cases as the model is, so to say, attacked by an adversarial attack outside of the distribution of the original data.
In some cases, we can use such a method to test the model on data shifts, as, for example, the attributions can shift heavily.
However, for our focus on the model itself, such an adversarial attack is interesting but does not show internal decision makings for the dataset we are interested in. 

\begin{figure}[h!tb]
    \centering
    \includegraphics[width=\textwidth,trim={0 11mm 0 0mm},clip]{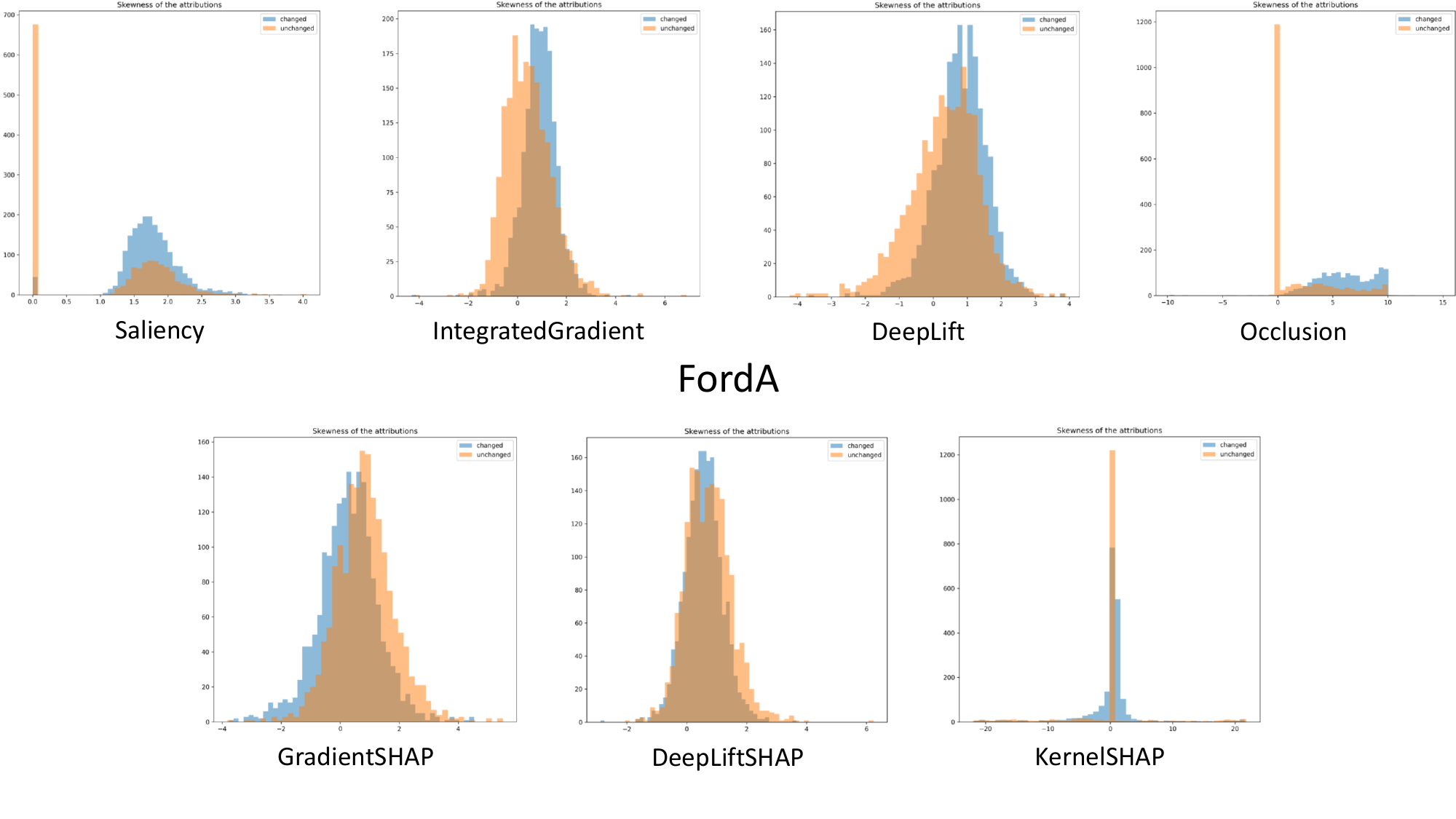}
    \caption{Skewness distribution of the attribution techniques on the FordA dataset. Clear differences in the distributions of the attributions are visible. Further, the differences between changed and unchanged sample predictions and their attributions in the distributions are observable and show two different patterns in the techniques.}
    \label{fig:skewness_dist}
    \vspace{-1em}
\end{figure}

However, we also notice that the perturbation strategy heavily influences the best working method.
If we switch, for example, to a perturbation to zero, we see Occlusion~\cite{zeiler_visualizing_2014} as the winner in~\autoref{fig:forda_results}.
Such a change in the best working technique demonstrates that the perturbation analysis just with one strategy is not enough to compare attribution techniques.
We need multiple strategies to decide on one technique.
However, we can also further take a deeper look into the attributions themselves.
Focusing on the different skewness of the attributions and their distributions as seen in~\autoref{fig:skewness_dist}, we can already see some trends toward techniques enabling an easier inspection of the method and how well the method performs for the perturbation analysis.
Especially, KernelSHAP in~\autoref{fig:skewness_dist} demonstrates a nice pattern with two nearly not overlapping distributions.
Such a nice distribution can help us to decide on one attribution technique.

\begin{figure}[h!tb]
    \centering
    \includegraphics[width=\textwidth,trim={0 21mm 0 1mm},clip]{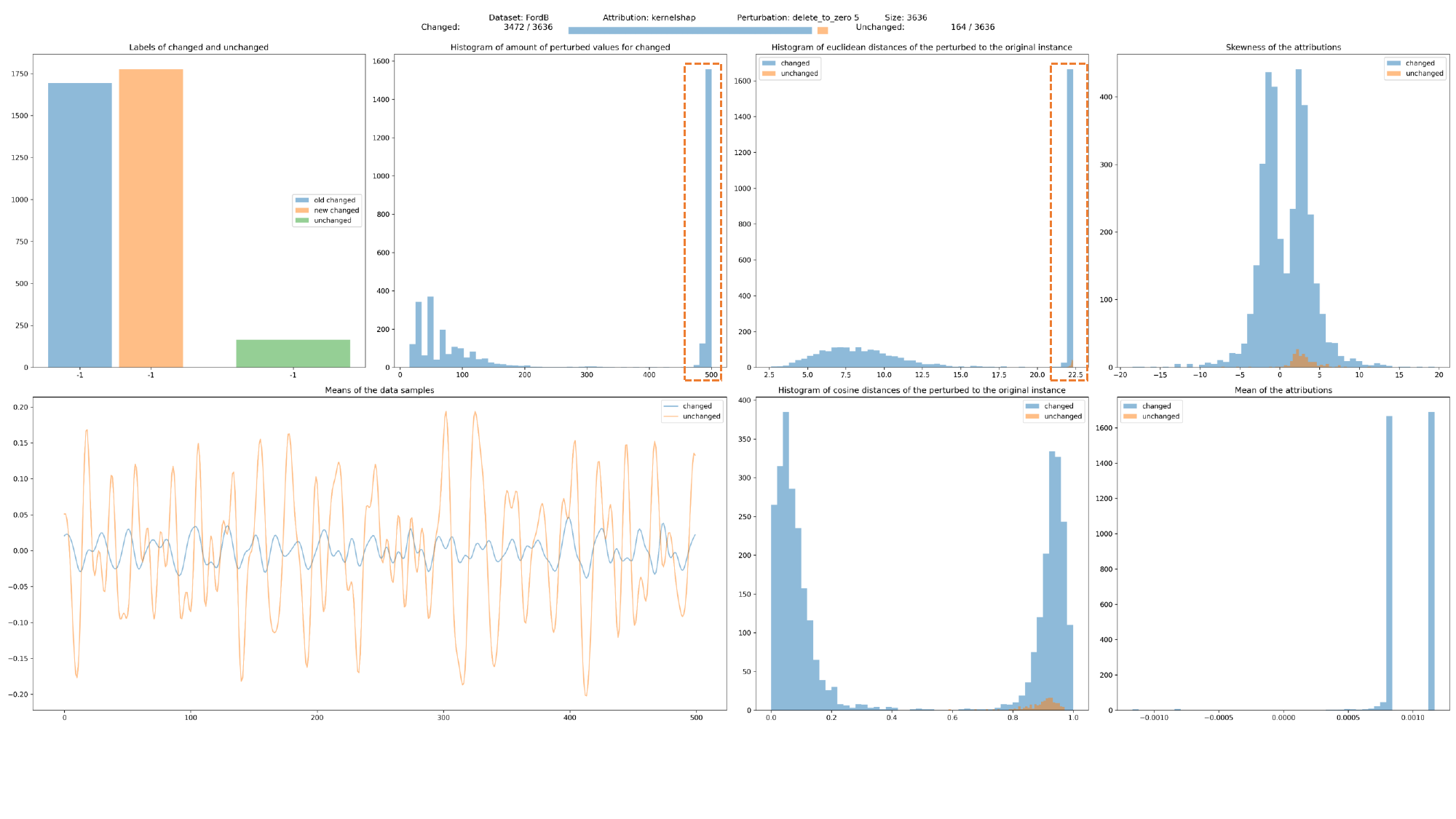}
    \caption{Perturbation analysis card for the FordB dataset. The visualizations show a very distinct pattern. For the means of the raw samples (R), the few unchanged samples compose quite a diverse pattern, while the changed ones go to zero based on their standardization. However, the plot on the top (C) with the orange marker presents a pattern we do not want to have in a perturbation analysis, as it shows that we need to perturb a lot of data to achieve a change in prediction.}
    \label{fig:perturbation_card_fordb}
    \vspace{-1em}
\end{figure}

\begin{figure}[!htb]
    \centering
    \includegraphics[width=0.99\textwidth,trim={0 43mm 77mm 0},clip]{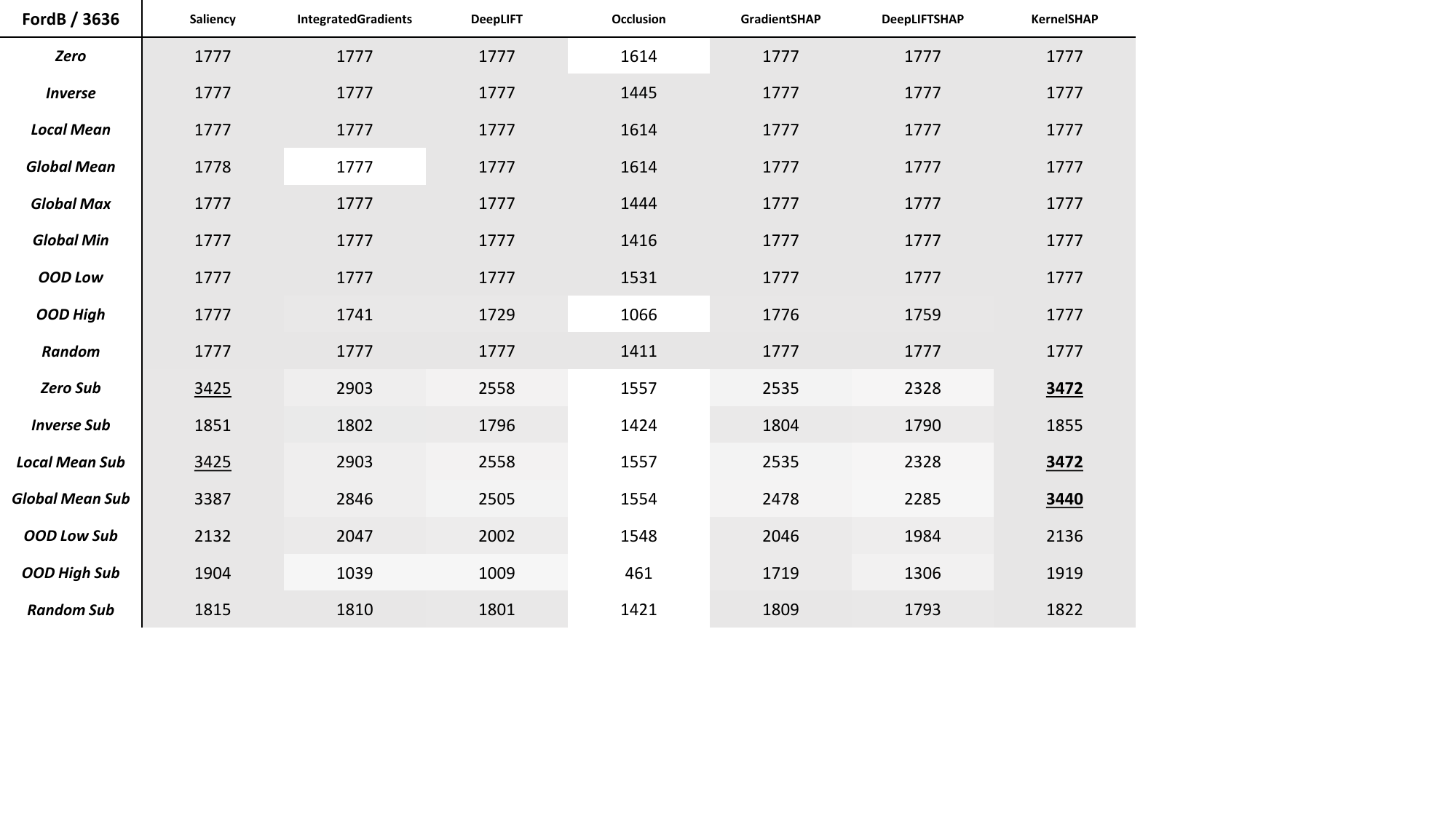}
    \caption{Changed samples from the perturbation analysis for the FordB dataset. The higher, the better. KernelSHAP is the winner, and Saliency is behind.}
    \label{fig:fordb_results}
\end{figure}

The model for the FordB dataset is a bit worse than for the FordA dataset, which leads, in most cases, to worse performance in the perturbation analysis~\cite{simic_perturbation_2022}.
However, again the KernelSHAP and Saliency generate good working attributions for the change in the prediction for the perturbation to zero strategy.
For this dataset, KernelSHAP achieves to change of 3472 from 3636 samples as seen in~\autoref{fig:fordb_results}.
Especially interesting is the distribution of the skewness of the attributions.
A more in detail analysis of these two peaks could lead to further insights into the model and the attributions, but such an analysis needs another visualization, e.g., projecting the attributions in a scatter plot.
However, if we further inspect our corresponding model card in~\autoref{fig:perturbation_card_fordb}, we can see that on the plot on the right, the change happens if a lot of values are removed from the original sample.
Such a result is also observable in the other perturbation cards for the other techniques.
In our study, we have identified a possible shortcut~\cite{geirhos_shortcut_2020} that our model has learned from the training data.
We speculate that the shortcut consists of a certain range or specific time points which need to be in a certain range of values to be classified as one class or the other class, and if we destroy this property, we change the prediction.
So, our model learns a static version or range for one class and classifies everything else into the other class.
Such a model does have more in common with an outlier detector than with a wanted classifier.
Thus, we identified a shortcut of the model to be able to improve the classification without using all available features~\cite{geirhos_shortcut_2020}.

\begin{figure}[h!tb]
    \centering
    \includegraphics[width=\textwidth,trim={0 21mm 0 1mm},clip]{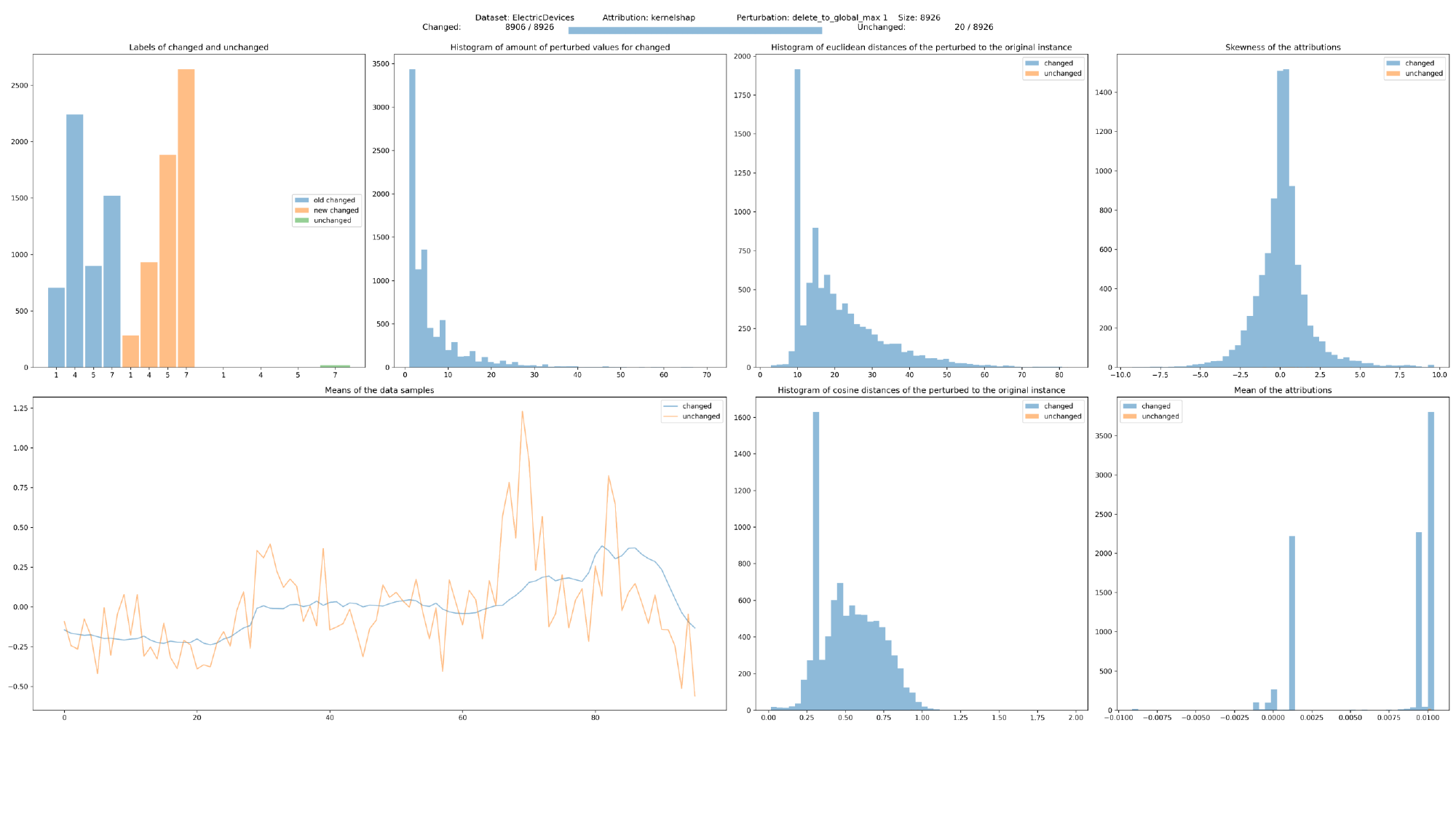}
    \caption{Perturbation analysis card for the ElectircDevices dataset. The skewness distribution is quite interesting as it nearly presents a Gaussian distribution, with the mean being more sparse and quite focused on only three large pillars.}
    \label{fig:perturbation_card_electricdevices}
    \vspace{-1em}
\end{figure}

The ElectricDevices dataset is harder for the model as we do not only have a binary classification problem but seven classes the model needs to separate.
However, as before, not even the state-of-the-art performance is as accurate as possible~\cite{ismailfawaz_deep_2019}, which leads to worse attributions and a more diverse perturbation analysis result.
Again, KernelSHAP performs best with a change of 8906 from 8926 samples with the values perturbed to the global max as seen in the perturbation card of~\autoref{fig:perturbation_card_electricdevices}.
However, also IntegratedGradients~\cite{sundararajan_axiomatic_2017} works well, but only with another perturbation strategy, namely changing the perturbed value to the global mean of the dataset.
The dataset demonstrates quite nicely that the attribution techniques need different perturbation strategies to reveal the models' internal decision-making.
Some of the techniques focus on different features the model learned as the ranking of the best-performing attribution techniques based on the perturbation analysis changes from strategy to strategy for this dataset.
Additionally, when we delve into the labels of the changed and unchanged predictions, we notice that various attribution methods alter different labels in the perturbation. 
For example, KernelSHAP seems to modify every class besides seven, whereas Saliency influences classes other than five and six.
However, unlike the previous FordB dataset, we do not see an unwanted perturbation pattern in the amount of perturbed values visualization. 
Such an effect presents that the attribution techniques are more suitable for the dataset and model than for the FordB model.

\begin{figure}[!htb]
    \centering
    \includegraphics[width=0.99\textwidth,trim={0 43mm 77mm 0},clip]{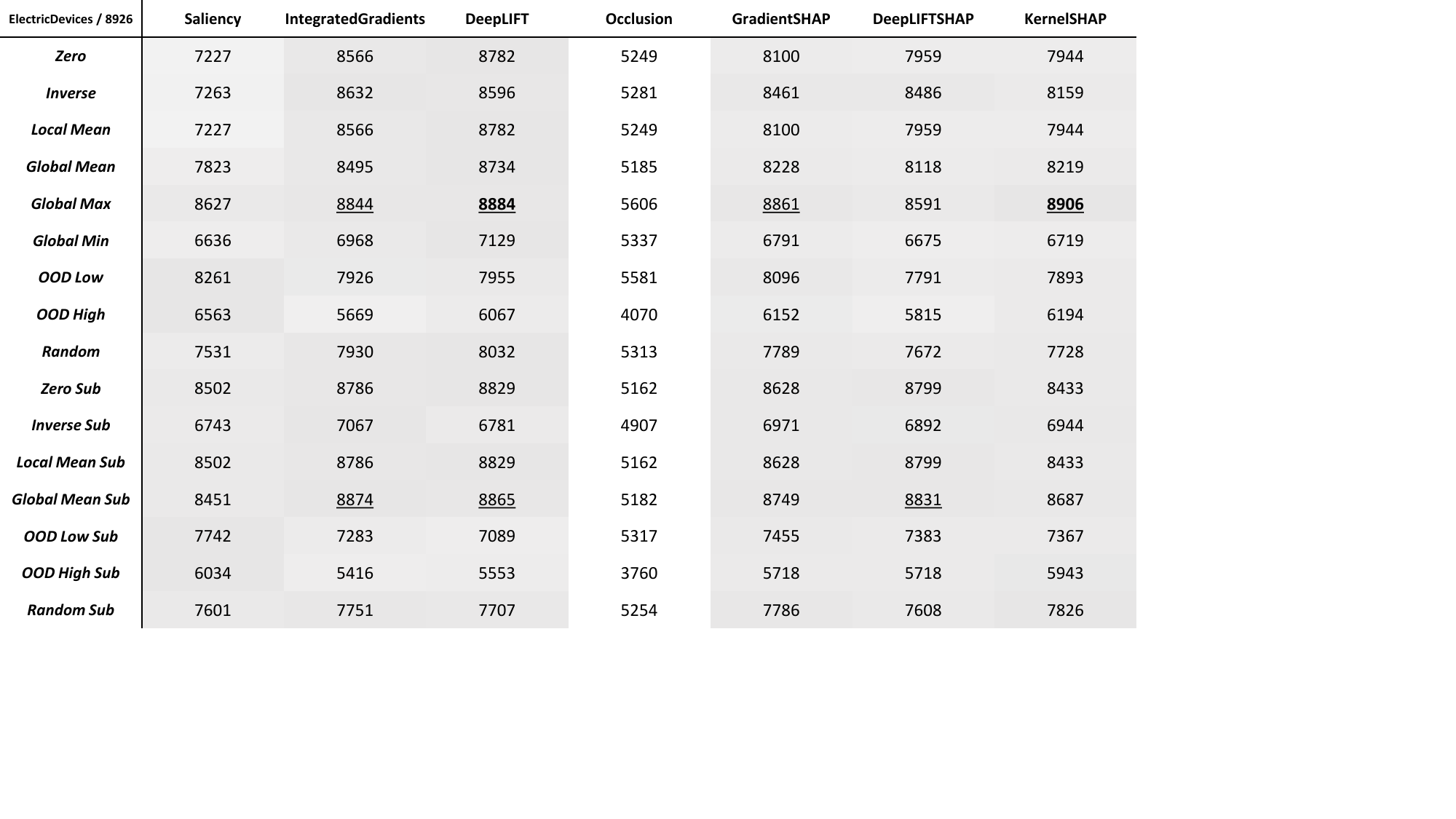}
    \caption{Changed samples from the perturbation analysis for the FordB dataset. The higher, the better. KernelSHAP is the winner, and DeepLIFT and IntegratedGradients are behind on different perturbation strategies.}
    \label{fig:electricdevices_results}
\end{figure}

\textbf{Summary --} As we have seen in our results (\autoref{fig:forda_results},~\autoref{fig:fordb_results},~\autoref{fig:electricdevices_results}), KernelSHAP performs quite well but takes a lot of time to calculate the attributions. 
Due to the sampling-based approach of KernelSHAP, the attributions are not deterministic and can vary from multiple computational runs.
Further, in many cases, Saliency (or Vanilla Gradient multiplied by the  Input) works surprisingly well and is only sometimes improved by additional extensions on top, such as IntegratedGradients.
Thus, Saliency provides a promising variant for future experiments and techniques on top of it.
So, if the attribution (explanation) is time-critical, Saliency is a well-suited method.
If it is not time-critical, KernelSHAP provides the best-working attributions based on our experiments.
Our collected data has even more insights and findings using the proposed perturbation analysis cards, which we look forward to analyzing and publishing with the code.
The published source code can be used as a framework to experiment on more datasets, and the perturbation analysis cards can be used to report the results.
The GitHub repository can be explored with more perturbation analysis cards and JSON data for the collected results of our experiments.

\section{Conclusion and Future Work}

After reviewing related work, we presented an in-depth analysis of perturbation strategies for attributions on time series.
With the analysis, we dug into a CNN trained on time series classification to investigate attributions, perturbation strategies, and shortcuts the network learned.
We presented our results in perturbation analysis cards to enable users to analyze the results in detail by inspecting the aggregated data in visualizations and comparing them easily with other techniques based on the provided cards.
We identified SHAP as a suitable method to generate working attributions in all experimented datasets.
Other gradient-based methods also work quite well but do not perform as well as, e.g., KernelSHAP.
However, depending on the perturbation strategy, the best working attribution technique changes quite drastically also for some techniques.
We advise not only focusing on a single strategy but to using multiple strategies and aggregating the results of these, and looking at the distribution of the skewness to enhance the comparability.
In our experiments, we also found a shortcut or spurious correlation for the FordB dataset, which our model learned to classify one class and to classify everything else as the other class.

\textbf{Future work -- }
We want to extend the experiment to other attribution techniques and compare the results with the already collected experiment results.
Also, we want to compare the attributions even in more detail by, e.g., aggregating the attributions and comparing them on a higher level to find matching patterns.
Different trends and subsequences are further patterns to analyze to gain knowledge into the attribution techniques.
With such an approach, we also want to include the \textit{local Lipschitz estimate}~\cite{alvarezmelis_robustness_2018} to rank consistent attributions higher.
Last, we want to extend the \textit{Perturbation Effect Size}~\cite{simic_perturbation_2022} and use our gained knowledge to combine perturbation strategies, switching predictions, and distances to generate a measure to evaluate attributions on time series classification models more robust and fully automatically to make it easier for users to decide which attributions to use for explanations.
We also want to enhance our perturbation analysis cards further to be more easily readable and comfortable for non-experts to be able to gain insights at a single glance.

\subsubsection*{Acknowledgements}
This work has been partially supported by the Federal Ministry of Education and Research (BMBF) in VIKING (13N16242).

%
%
%
\bibliographystyle{splncs04}
\bibliography{main}

\end{document}